\documentclass{article}

\usepackage[preprint]{AderseralRobust}

\usepackage[utf8]{inputenc} 
\usepackage[T1]{fontenc}    
\usepackage{hyperref}       
\usepackage{url}            
\usepackage{booktabs}       
\usepackage{amsfonts}       
\usepackage{amsmath}
\usepackage{nicefrac}       
\usepackage{microtype}      
\usepackage{multirow}
\usepackage{graphicx}
\usepackage{comment}
\usepackage{wrapfig}

\graphicspath{ {./} }

\usepackage{xcolor}

\newcommand{\dvg}{KL}

\title{Understanding Adversarial Behavior of DNNs by Disentangling Non-Robust and Robust Components in Performance Metric}

\author{%
  Yujun Shi \thanks{Work done when Yujun Shi was interning in Tencent.} \\
  Nankai University \\
  \texttt{1511201@mail.nankai.edu.cn} \\
  \And
  Benben Liao \thanks{Benben Liao has made the equal contribution to this work.}\\
  Tencent \\
  \texttt{bliao@tencent.com}
  \And
  Guangyong Chen \thanks{Corresponding author.}\\
  Tencent \\
  \texttt{gycchen@tencent.com}
  \And
  Yun Liu \\
  Nankai University \\
  \texttt{nk12csly@mail.nankai.edu.cn}
  \And
  Ming-Ming Cheng \\
  Nankai University \\
  \texttt{cmm@nankai.edu.cn} \\
  \And
  Jiashi Feng  \\
  National University of Singapore \\
  \texttt{elefjia@nus.edu.sg} \\
}

\begin{document}

\maketitle

\begin{abstract}
The vulnerability to slight input perturbations is a worrying yet intriguing property of deep neural networks (DNNs). Despite many previous works studying the reason behind such adversarial behavior, the relationship between the generalization performance and adversarial behavior of DNNs is still little understood. In this work, we reveal such relation by introducing a metric characterizing the generalization performance of a DNN. The metric can be disentangled into an information-theoretic non-robust component, responsible for adversarial behavior, and a robust component. Then, we show by experiments that current DNNs rely heavily on optimizing the non-robust component in achieving decent performance. We also demonstrate that current state-of-the-art adversarial training algorithms indeed try to robustify the DNNs by preventing them from using the non-robust component to distinguish samples from different categories. Also, based on our findings, we take a step forward and point out the possible direction for achieving decent standard performance and adversarial robustness simultaneously. We believe that our theory could further inspire the community to make more interesting discoveries about the relationship between standard generalization and adversarial generalization of deep learning models.
\end{abstract}

\section{Introduction}
Deep neural networks (DNNs) have achieved enormous success in many different tasks \cite{lecun2015deep} over the last decade. There is a major line of works trying to boost the performance of deep learning models from different aspects \cite{krizhevsky2012imagenet,simonyan2014very,szegedy2015going,he2016deep,he2015delving,ioffe2015batch}. While the community is devoting to achieve new state-of-the-art performance with DNNs, some researchers \cite{szegedy2013intriguing} identify these powerful models are susceptible to perturbations that are even imperceptible by human. Consequently, a solid body of works have been developed on both finding the most effective attacks \cite{goodfellow2014explaining,moosavi2016deepfool,kurakin2016adversarial,zhao2018adversarial} as well as obtaining relatively more adversarial robust models \cite{goodfellow2014explaining,miyato2015distributional,madry2017towards,zhang2019theoretically}. Specifically, in \cite{zhao2018adversarial}, the authors provide theoretic derivation on how Fisher information of the model's output w.r.t input characterizes the adversarial behavior around input.

Exploiting the model's ability to stay adversarial robust \cite{miyato2015distributional,madry2017towards} also reveals a non-trivial degradation in standard performance as a cost of being relatively more adversarial robust, which has attracted much attention recently. Some works \cite{schmidt2018adversarially,tsipras2018robustness,stutz2018disentangling,nakkiran2019adversarial,adversarialfeatures2019} have emerged to try to understand this apparently trade-off between standard generalization and adversarial generalization. For example, in \cite{adversarialfeatures2019}, the authors provide a novel view point that adversarial samples are non-robust features that could help the generalization of deep learning models and validate their conjecture by experiments.

In this work, however, we theoretically reveal the relationship between standard performance as well as adversarial behavior of deep learning models with Taylor expansion of Kullback–Leibler divergence (KL divergence) and Fisher information. Interestingly, we show that the overall performance objective could be disentangled into a non-robust component, which has the side effects of causing adversarial behavior, as well as a robust component. Our analysis shows that it is indeed the fact that adversarial robustness and high standard performance are contradictory as the non-robust component, which contribute to the standard performance also cause adversarial behavior.

The accuracy on test set is usually used to measure the performance of machine learning models for classification tasks. In this work, to investigate the adversarial behavior of DNNs, we propose to transform such performance metric into the KL divergence between output distributions of test set samples from different categories. This new objective could not only perfectly characterize how well the model distinguishes samples from different categories, but also connect model performance and its adversarial behavior tightly. The developed theory conveys that the overall performance objective could be disentangled into a non-robust component, which has the side effects of causing adversarial behavior, as well as a robust component.

We also demonstrate by experiments that current deep learning models rely heavily on the non-robust component to generalize, which is the underlying reason for adversarial behavior in trained deep learning models. In addition, we show that state-of-the-art adversarial training algorithms are all trying to constrain the model from using the non-robust components. Based on the above findings, we suggest that there might exist a perfect balance point for the deep learning models to possess decent standard generalization ability while stay adversarial robust.

Our contribution is summarized as follows.
\begin{itemize}
    \item We propose a new metric that could both characterize the standard performance and better connect with adversarial behavior of DNNs.
    \item By properly expanding our metric with Fisher information, we quantitatively explain the relationship between standard performance and adversarial robustness of DNNs.
    \item We then take a step forward and point out possible direction for achieving decent standard performance and adversarial robustness simultaneously.
\end{itemize}


\section{Related Work}
\paragraph{Adversarial Attacks and Defense}
To study the adversarial behavior of deep learning models, many algorithms in terms of both finding the most effective adversarial perturbation and improving adversarial robustness of the deep learning system have been proposed recently. \cite{goodfellow2014explaining} proposes an one-step attack algorithm based on gradient of the deep learning models called fast gradient sign method (FGSM), \cite{papernot2016limitations} propose an attack algorithm based on jacobian saliency map, \cite{moosavi2016deepfool} proposes an attack algorithm based on Newton's iterative algorithm for finding roots of a non-linear function. \cite{papernot2016distillation} proposes a training mechanism for defense based on distillation while \cite{carlini2017towards} proposes the "CW attack" that could render the distillation defense mechanism useless. \cite{kurakin2016adversarial} proposes an iterative version of the FGSM attack, and \cite{madry2017towards} suggests that the adversarial training based on projected gradient descent attack algorithm has universal defense effects. \cite{miyato2015distributional} proposes the distribution smoothing training strategy for defense in both supervised and semi-supervised learning setting. \cite{zhao2018adversarial} studies the adversarial attack under the Fisher information metric and uses the power method to solve eigenvectors of the Fisher information matrix, where the eigenvector of the greatest eigenvalue is treated as the attack noise.
\vspace{-2mm}
\paragraph{Fisher Information in Deep Learning}
\cite{chaudhry2018riemannian} uses Fisher information matrix as a metric that induces distance on parameter manifold and develops an regularization term for incremental learning. The concept of Fisher information is also widely applied in deep learning algorithm based on natural gradient descent \cite{pascanu2013revisiting,desjardins2015natural}, as well as meta-learning~\cite{andrychowicz2016learning,achille2019task2vec}. In these works, they usually view the parameter of the model as the parameter of the Fisher information matrix. Recently, \cite{miyato2015distributional,zhao2018adversarial} propose to use Fisher information to represent the local geometry of the log-likelihood landscape of the deep learning model so that the adversarial behavior could be better studied. In these two works, however, they view the input of the deep learning model as Fisher information's parameter instead. We also adopt the same definition for Fisher information as \cite{miyato2015distributional,zhao2018adversarial} to better study the relationship between the standard performance and adversarial robustness of the model.
\vspace{-2mm}
\paragraph{Adversarial Behavior of DNNs}
Previously, there are a line of works trying to explain the robust model's degradation in standard performance~\cite{nakkiran2019adversarial,stutz2018disentangling,tsipras2018robustness,schmidt2018adversarially,zhang2019theoretically}. In these work, they all start by studying the accuracy of the model, and sometimes with relatively strong assumptions. The recent literature \cite{adversarialfeatures2019} also empirically shows that adversarial samples are features that help the standard generalization of deep learning models by disentangling non-robust and robust features. In our work, we theoretically point out that it is indeed the fact that what causes adversarial behavior also helps boost the model's performance with little assumptions. Our theory, however, doesn't indicate that achieving decent performance and being adversarial robust are two completely contradict objectives. It is still possible that there exist a perfect balance point where both objectives could be satisfied simultaneously.

\section{Disentanglement of the Performance Metric}
We present our main results in this section. We first propose to transform the standard testing-accuracy based performance metric to a KL divergence based one. Then with the newly formed objective, we derive the performance of a DNN model is indeed determined by  two disentangled components and  it is one of them that introduces the  model adversarial behavior. We finally conclude this section by presenting some complementary understandings from the viewpoint of information geometry.

\subsection{Proposed performance metric}
Prediction accuracy on test set is typically adopted to measure   performance of a DNN model in    the  literature. However,   such   metric indeed hides  the connection between the model performance and  its adversarial behavior. In this work,  in order to build more transparent connection and better understand model  adversarial behavior, we  propose to adopt  the average KL divergence between output distribution of any pair of data of different categories as the classification performance metric. 

We  denote $x$ as the input image and $y$ as the corresponding one-hot label distribution, $f$ as the model and $\hat{y} = f(x)$ as the output distribution of the model, $N_{pair}$ as the number of pairs of input data with different labels, $JS(f(x_i)\|f(x_j))$ as the Jensen-Shannon divergence  between $f(x_i)$ and $f(x_j)$. We propose to adopt the objective as \emph{Cross Category KL Divergence} (CCKL):
\begin{equation}
    \label{eq: 1}
    CCKL = \frac{1}{2 N_{pair}}\sum_{\forall y_i \neq y_j} KL(f(x_i)||f(x_j)).
\end{equation}
By Lin's inequality \cite{lin1991divergence},
we can derive the following lower bound from triangular inequality for describing the relation between the widely used cross entropy loss and our proposed objective  \eqref{eq: 1}:
\begin{equation}
    \label{eq: 2}
    KL(f(x_i)\|f(x_j))\geq 2 JS(y_i\|y_j)-KL(y_i\|f(x_i))-KL(y_j\|f(x_j)), \forall y_i \neq y_j,
\end{equation}
where $JS(y_i\|y_j)$ denotes the Jensen-Shannon divergence between $y_i$ and $y_j$.
This lower bound of $KL(f(x_i)\|f(x_j))$ effectively characterizes its behavior along the training process. As the lower bound indicates, the training error $KL(y_i\|f(x_i))$ decreases so the whole lower bound increases during training process.

\begin{figure}
  \centering
  \includegraphics[width=10cm]{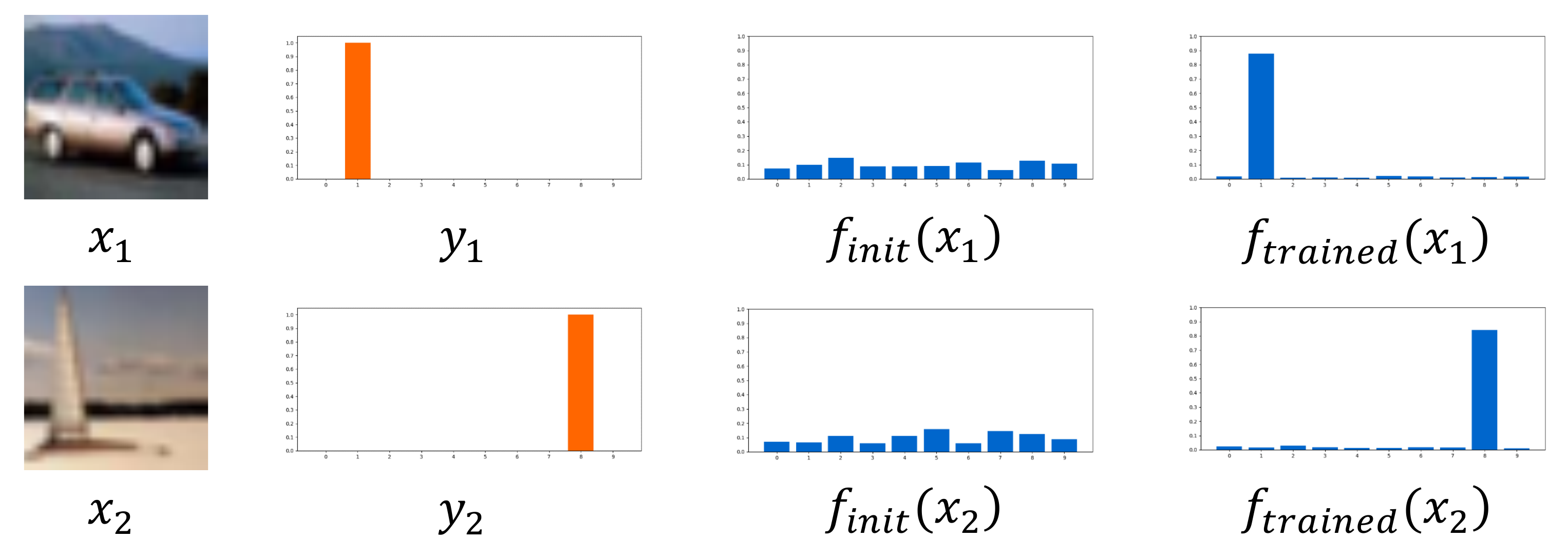}
  \caption{Graphical illustration of how $KL(f(x_1)\|f(x_2))$ evolves during training, (left) ground-truth label, (middle) network outputs when initialized, (right) network outputs when optimum reached.}
  \label{fig_performance_objective}
\end{figure}

We can also view this from another perspective. When randomly initialized, a DNN $f$ does not have any knowledge for classifying samples correctly. Therefore, it does not distinguish different inputs $x_i,x_j$, and the output distributions shall be similar, as shown in Figure \ref{fig_performance_objective}. That is to say, $KL(f(x_i)\|f(x_j))$ shall be relatively small at the very beginning of the training stage. As training proceeds, more  label dependent information is integrated into the model $f_\theta$. The network becomes better for generalization and the output distribution $f_\theta(x)$ on test set gets closer to the true label distribution $y$. At late training stage, the model loss $KL(y\|f(x))$ on average will decrease to relatively small value. 
By continuity of KL divergence, $KL(f(x_i)\|f(x_j))$ will be sufficiently close to $KL(y_i\|y_j)$ 
as shown in Figure \ref{fig_performance_objective}.

\begin{figure}
  \includegraphics[width=4.5625cm]{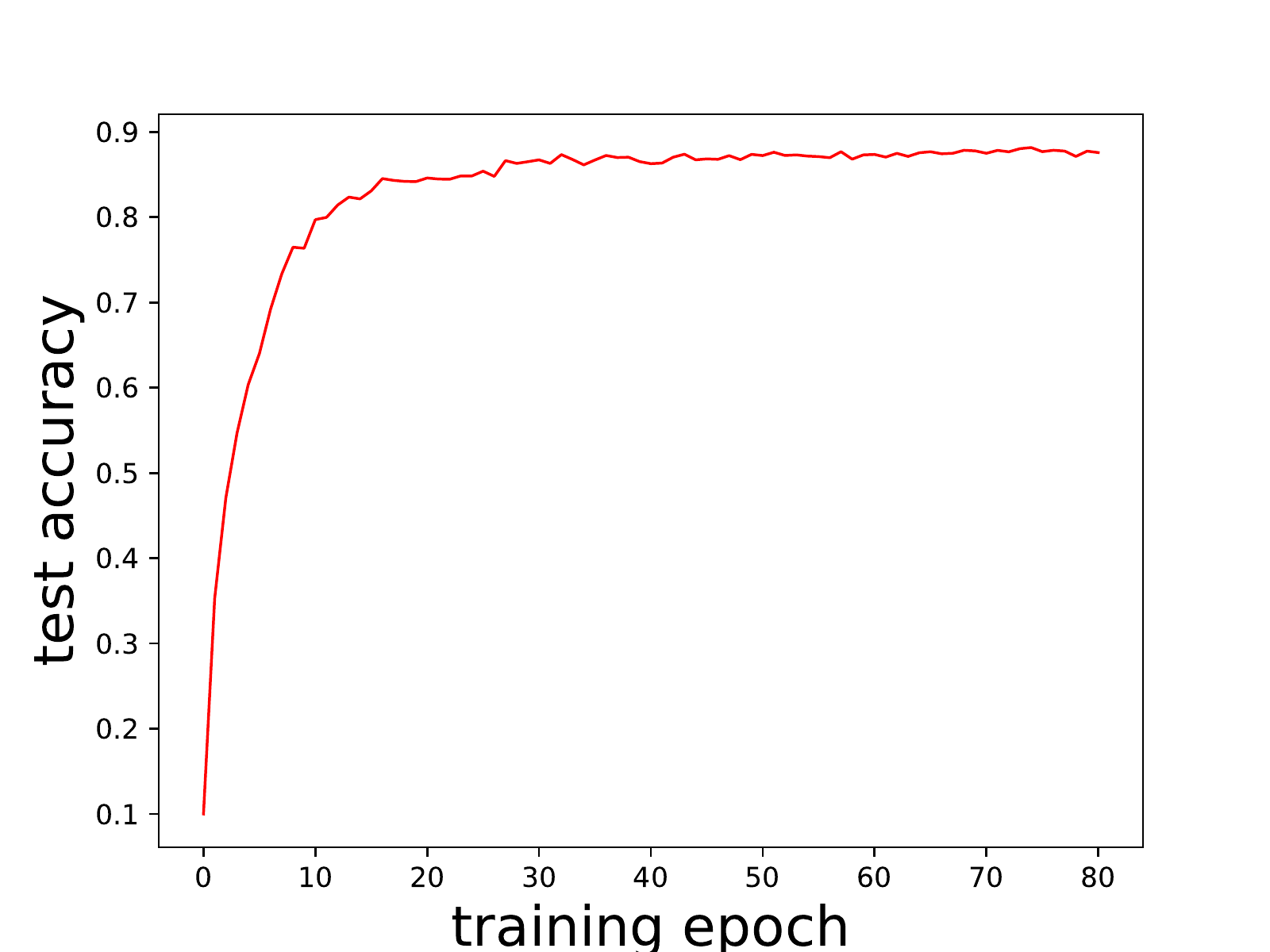}
  \includegraphics[width=4.5625cm]{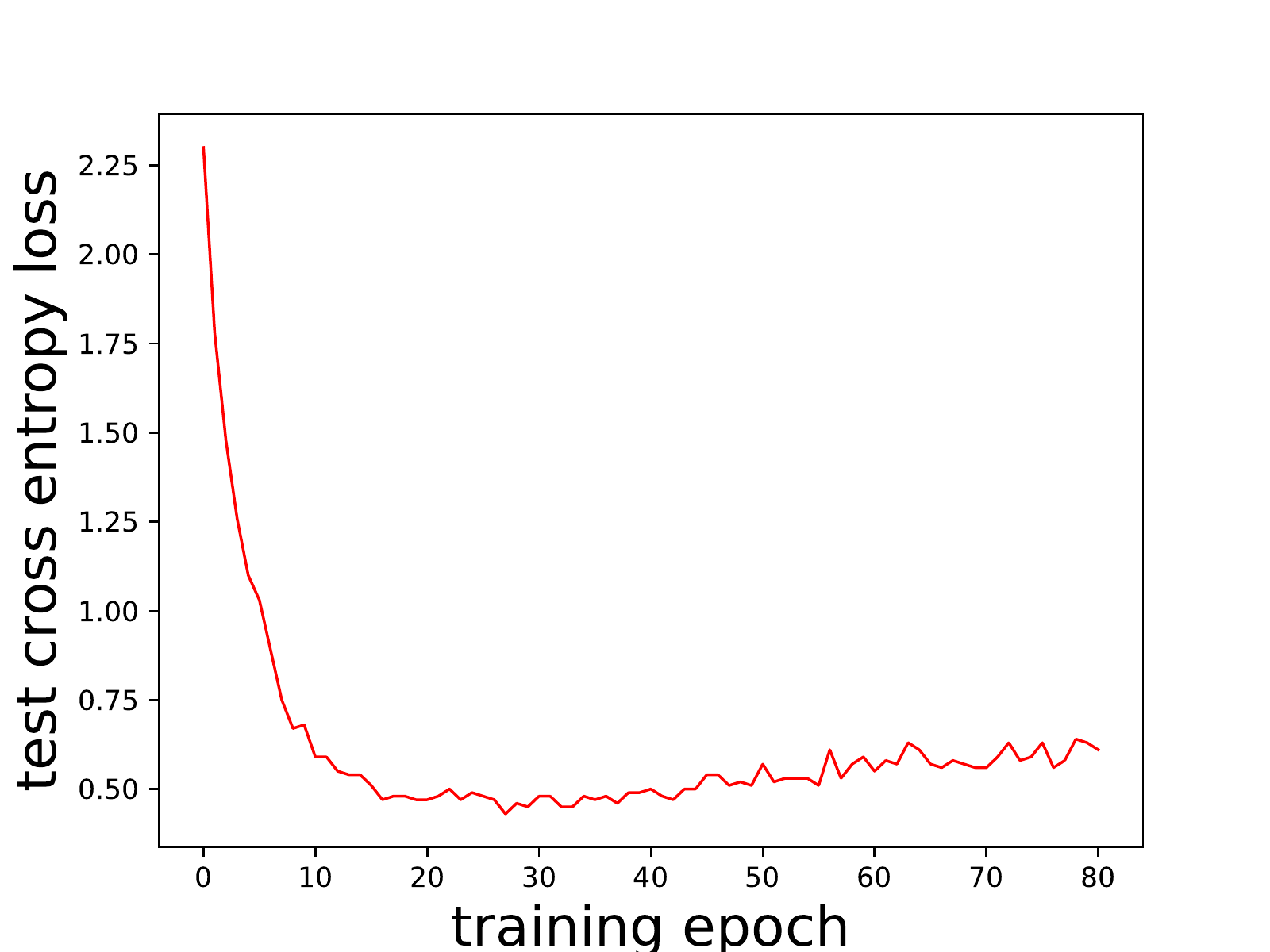}
  \includegraphics[width=4.5625cm]{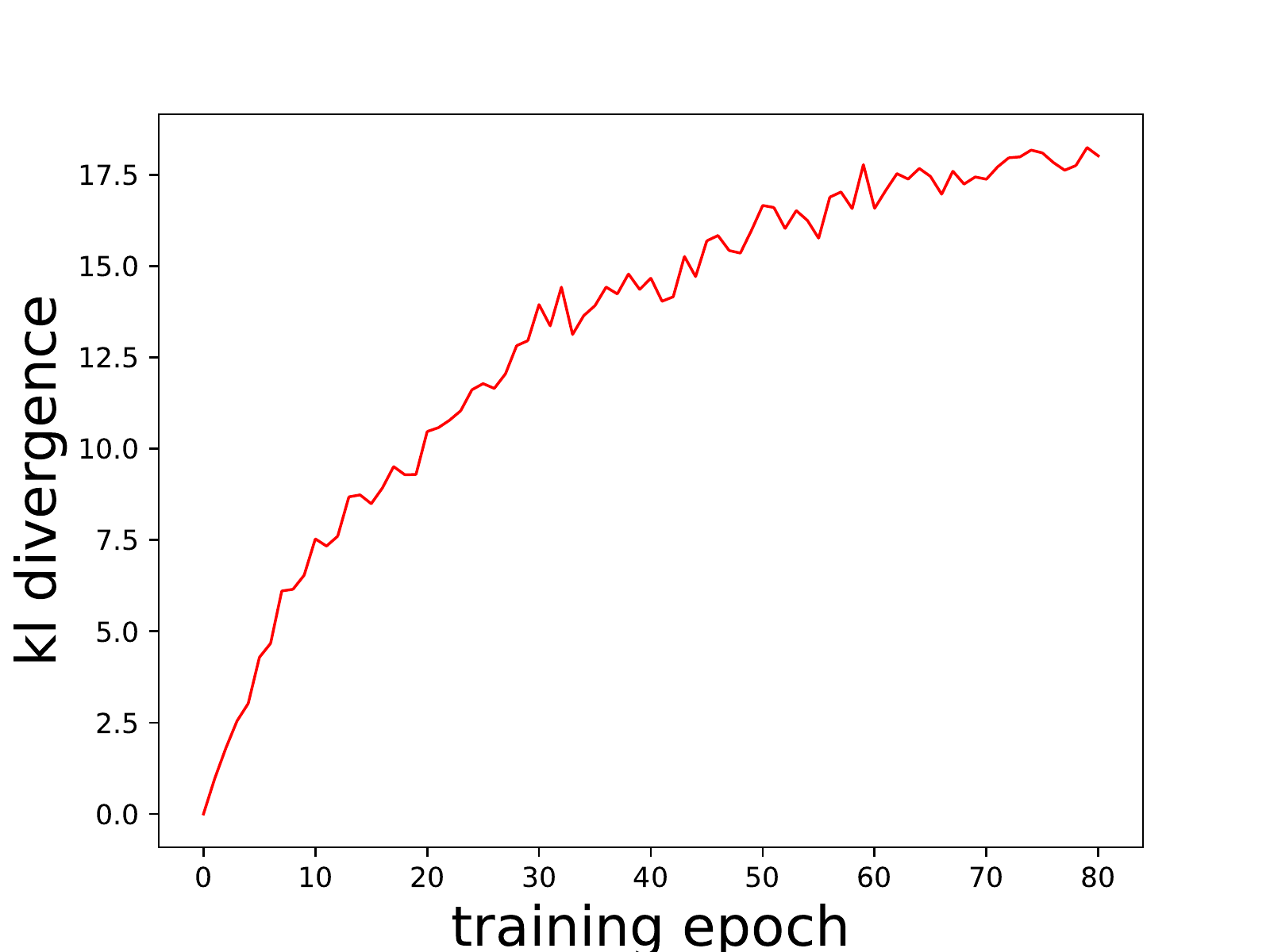}
  \caption{Visualization of how (left) accuracy, (middle) cross entropy loss and (right) CCKL on test set evolve when training  VGG13  on CIFAR-10. It is clear that CCKL well indicates    test accuracy and cross-entropy training loss. }
  \label{fig_acc_kl}
\end{figure}

This  proposed measure could better characterize how well the model distinguish the input data from data of other categories. To further expound on our point, we provide a visualization about this relationship in Figure \ref{fig_acc_kl}. 

\subsection{Connections Between Adversarial Behavior and CCKL}
We now explain how the above CCKL objective could be connected to the model's adversarial behavior. In most previous literature \cite{goodfellow2014explaining,kurakin2016adversarial,madry2017towards}, the following cross entropy loss of the model $f_\theta$ on adversarial samples (with perturbation $\eta$)
$$\min_\theta \max_\eta KL(y\|f_\theta(x+\eta))$$
is adopted to study adversarial behavior of the DNNs. 
As training proceeds and the parameter $\theta$ approaches the optimum, the output distribution $f_\theta(x)$ will be close to true label $y$. In this situation, since KL divergence is continuous in the first variable so we have
$$KL(y\|f_\theta(x+\eta))\approx KL(f_\theta (x)\|f_\theta(x+\eta)).$$
This observation inspires us to
study the model adversarial behavior  from a distribution point of view, as explored in \cite{miyato2015distributional,zhao2018adversarial,zhang2019theoretically}. Instead of viewing the output of $f_\theta$ as a single scalar, we treat the model $f_\theta$ as a function that outputs a prediction distribution over the input. Thus we use the KL divergence between the output distribution over the original samples and adversarial samples to characterize  the model's adversarial behavior.

In order to better connect adversarial behavior with the CCKL proposed before, we define the following adversarial measure built upon the above relation:
\begin{equation}
    \label{eq: 3}
    L_\theta(x, \eta) = \dvg(f_\theta(x)\|f_\theta(x+\eta)).
\end{equation}
The objective of the corresponding adversarial training is thus formulated as follow:
\begin{equation}
    \label{eq: 4}
    \min_\theta \max_{\eta} L_\theta(x, \eta) \;\; \text
    {s.t.}\; \|\eta\| \leq \epsilon.
\end{equation}
Based on the distribution point of view, \cite{zhao2018adversarial} reports state-of-the-art results in the task of adversarial attack and \cite{zhang2019theoretically} reports the state-of-the-art results in adversarial robustness. Therefore, using \eqref{eq: 3} instead of the cross entropy loss on adversarial samples to represent adversarial behavior is further justified.

Given the definition of $L_\theta(x, \eta)$, applying Taylor expansion yields the following:
\begin{equation}
    \label{eq: 8}
    \max_{\eta} L_\theta(x, \eta) = \max_{\eta} \eta^\top F_{x}\eta + \sum_{k=3}^{\infty} T_{x}^{(k)}(\eta) \;\; \text{s.t.}\; \|\eta\| \leq \epsilon.
\end{equation}
where $F_{x}$ is the Fisher information of $f(x)$ w.r.t.\ $x$. Let $f_j(x)$ be the $j$-th entry of $f(x)$ and $n$ be the number of entries of $f(x)$. Then $F_x$ can be calculated by:
\begin{equation}
    \label{eq: 7}
    F_{x}=\sum_{j=0}^n f_j(x) (\nabla_x \log f_j(x)) (\nabla_x \log f_j(x))^\top.
\end{equation}
When $\epsilon$ is sufficiently small, higher order terms in the above would vanish and \eqref{eq: 8} could be   simplified into:
\begin{equation}
    \label{eq: 9}
    \max_{\eta} L(x, \eta) = \max_{\eta} \eta^\top F_{x}\eta \;\; \text{s.t.}\; \|\eta\| \leq \epsilon.
\end{equation}
By setting $\nabla_{\eta}L = 0$, we obtain $F_{x}\eta = \lambda_{\max}\eta$, where $\lambda_{\max}$ is the maximum eigenvalue of $F_x$. Therefore, the solution to the overall adversarial objective corresponds to the leading eigenvector of $F_x$. Consequently, we have:
\begin{equation}
    \label{eq: 10}
    L(x, \eta) \leq \lambda_{\max} \epsilon^2 \;\; \text{s.t.}\; \|\eta\| \leq \epsilon.
\end{equation}
Note that $\lambda_{\max}$ here is also the spectral norm of the Fisher information matrix $F_{x}$. The above derivation shows that the local adversarial behavior of the model $f$ around input $x$ is determined by the spectral norm of Fisher information matrix: the adversarial behavior around $x$ would be more severe if the spectral norm of $F_x$ is larger.

Given two data-label pairs $(x_i, y_i)$ and $(x_j, y_j)$ with $y_i \neq y_j$, we could rewrite $KL(f(x_i)\|f(x_j))$ as:
\begin{equation}
    \label{eq: 11}
    \dvg(f(x_i)\|f(x_j)) = \dvg(f(x_i)\|f(x_i + (x_i - x_j))) = L(x_i, x_j - x_i).
\end{equation}
Therefore, we apply the same Taylor expansion as above and obtain:
\begin{equation}
    \label{eq: 12}
    \dvg(f(x_i)\|f(x_j)) = L(x_i, x_j-x_i) = (x_j-x_i)^\top F_{x_i}(x_j - x_i) + \sum_{k=3}^{\infty} T_{x_i}^{(k)}(x_j-x_i).
\end{equation}
Comparing \eqref{eq: 12} and \eqref{eq: 9}, we could further notice that they share the same Fisher information $F_{x_i}$. Therefore, the adversarial behavior at each data point and performance objective CCKL could be connected by Fisher information at each data point. 

{\bf Cram\'er-Rao bound} The adversarial training proposed in the paragraphs above constrains the input-output Fisher information of a DNN model. This constrain is a criteria of a good DNN model due to the following reasons. Recall the well-known Cram\'er-Rao bound 
$$\texttt{var}(\hat x)F_x\geq 1$$
says that if we try to use the output probability $f(x)$ to a statistics $\hat x$ to reconstruct the input $x$, the uncertainty in terms of variance $\texttt{var}(\hat x)$ is bounded below by the inverse of Fisher information $F_x$. For a DNN model that represents the reality, when it classifies an image with a correct label, say a dog, the label does not have any information about the environments - what color the dog is, where is the dog, adversarial perturbation, etc. Therefore, one cannot use the information contained in the label to reconstruct the original image. This means that the variance $var(\hat x)$ of any statistics $\hat x$ derived from output distribution $f(x)$ is relatively large for a good DNN model. In view of Cramer-Rao bound, this implies that the Fisher information of a DNN is a relatively small value.






\subsection{Disentanglement of the Performance Metric}
\label{section3_3}
In this section, we reveal the proposed performance objective, unifying measure of performance and robustness, could be decomposed into two components. To see this, we first denote $(x_j-x_i)^\top F_{x_i}(x_j-x_i)$ in \eqref{eq: 12} as $G_1$ and the following terms $\sum_{k=3}^{\infty} T_{x_i}^{(k)}(x_j-x_i)$ as $G_2$. Thus $KL(f(x_i)\|f(x_j))$ could be formulated as:
\begin{equation}
    \label{eq: KL_distentangle}
    KL(f(x_1)\|f(x_2)) = G_1 + G_2.
\end{equation}
Taking a closer look into \eqref{eq: KL_distentangle}, we could notice that the increase of $G_1$ and $G_2$ could both contribute to the rise of $KL(f(x_i)\|f(x_j))$, which is the performance objective. Note that since $G_1$ is a second order polynomial induced by $F_{x_i}$, and $(x_j-x_i)$ is fixed distance between two input $x_i$ and $x_j$, the rise of $G_1$ would asymptotically result in the rise of norm of $F_{x_i}$. That is to say, if the model rely heavily on the increase of $G_1$ to boost performance, the norm of $F_{x_i}$ would have to increase drastically. However, according to \eqref{eq: 10} and our derivation before, the rise in spectral norm of $F_{x_i}$ means more severe adversarial behavior around $x_i$. The trade off between standard performance and adversarial behavior is thus clearly characterized here: the model could rely on $G_1$ to boost performance, but it comes with the side effect of more severe adversarial behavior. What should also be noted, however, is that $G_2$ also contributes to the overall performance objective while not involves in the adversarial objective. Therefore, relying on terms in $G_2$ to distinguish $x_i$ from data belonging to other categories would not cause adversarial behavior. Therefore, we successfully disentangled the non-robust component and robust component in the overall performance objective.
\begin{wrapfigure}{r}{0.5\textwidth}
  \centering
  \includegraphics[width=6cm]{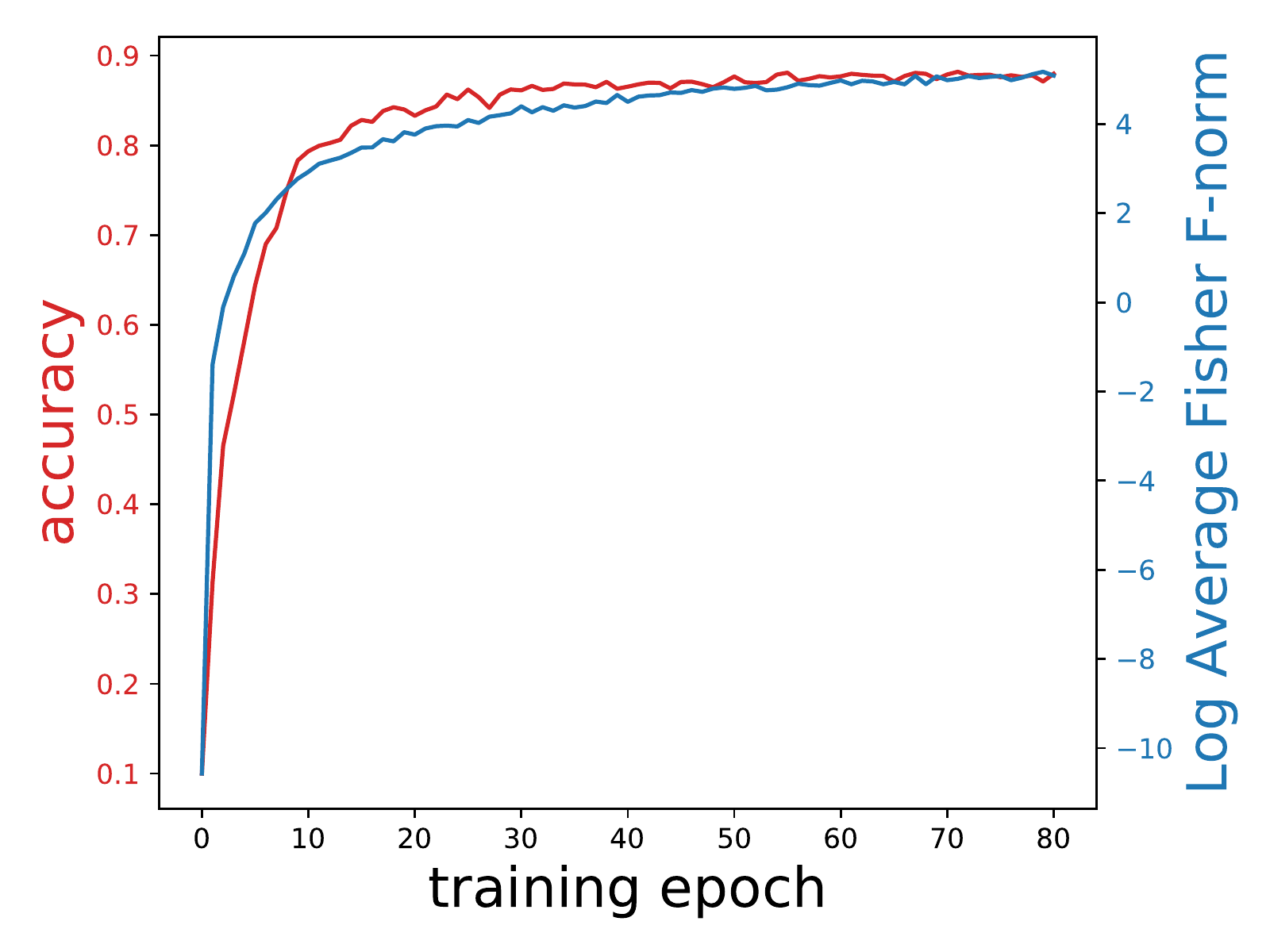}
  \caption{Visualization of how standard accuracy and average F-norm of Fisher information matrix on test set vary during training. The experiment is conducted on a VGG13 model and CIFAR-10 data set (for the same experiment for ResNet see section A of appendix). We take the logarithm to better visualize the average F-norm of Fisher information.}
  \label{fig_nat_acc_fisher}
  \vspace{-4mm}
\end{wrapfigure}

To further understand the role of $G_1$ in  classification, we visualize how F-norm of Fisher information evolves during training. Note that we visualize F-norm instead of spectral norm because all norms are equivalent and spectral norm is not computation feasible in our case. We first empirically show how the average F-norm of Fisher information on the test set and standard test accuracy vary during nature training process. The visualization is in Figure \ref{fig_nat_acc_fisher}. According to the statistics, we could observe that the norm of Fisher information increase drastically with the rise of accuracy, which indicates that current deep learning model rely heavily on the non-robust component $G_1$ to boost performance.

\begin{figure}
  \centering
  \includegraphics[width=6cm]{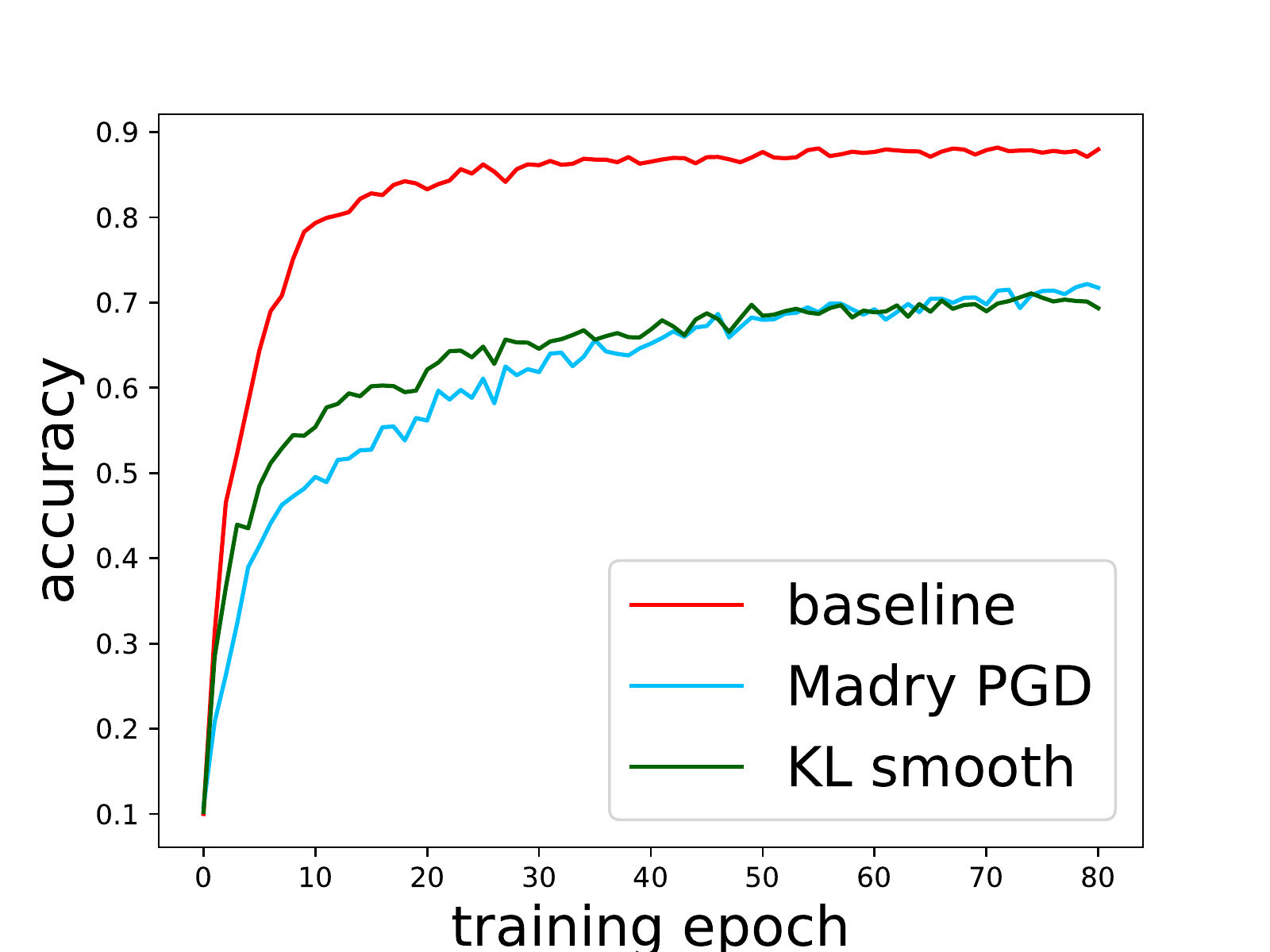}
  \includegraphics[width=6cm]{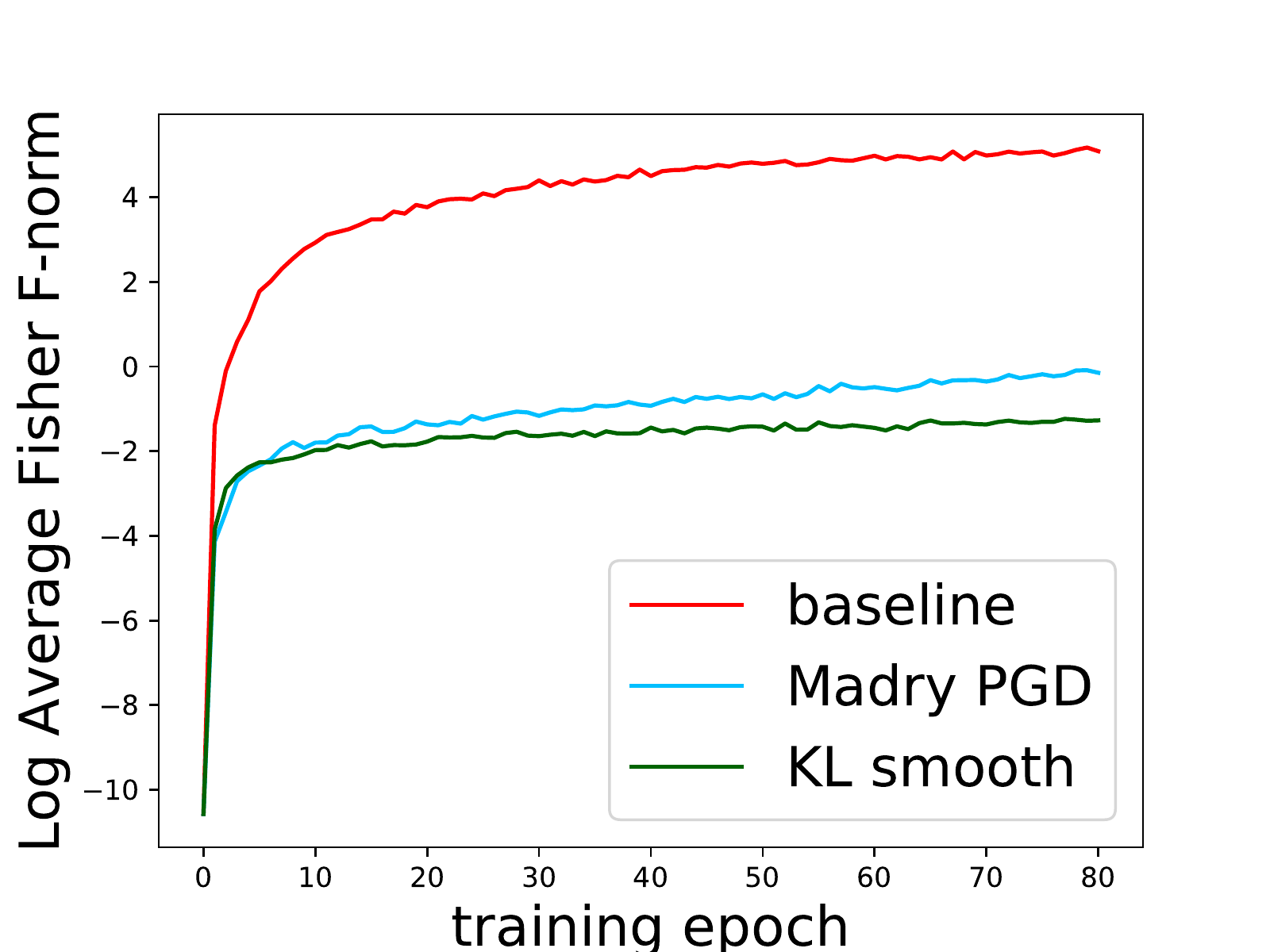}
  \caption{Visualization of how (left) standard test accuracy and (right) average F-norm of Fisher Information Matrix on test set vary during nature training and adversarial training with VGG13 on CIFAR-10 (for the same experiments for ResNet see section A of appendix). We take the logarithm to better visualize the average F-norm of Fisher information.}
  \label{fig_adv_acc_fisher}
\end{figure}

Then we compare nature training with the two state-of-the-art adversarial training algorithms \cite{madry2017towards,zhang2019theoretically} using the same visualization method. The result is shown in Figure \ref{fig_adv_acc_fisher}. It is clear that during adversarial training, although the Fisher information's average F-norm also rises with standard accuracy, the values of which is significantly smaller than its counterpart during nature training. That is to say, the adversarial training process could effectively constraint the model from using Fisher information in boosting performance.

Our experiments also demonstrate the widely known but little understood fact that the standard accuracy of models under the two adversarial training algorithms are significantly lower than their counterpart under nature training. According to our theory, it is because they are unable to effectively rely on adversary-prone non-robust components such as Fisher information to distinguish the input data from data belonging to other categories. Therefore, we theoretically and empirically demonstrate the relationship between standard accuracy and adversarial robustness using the disentanglement proposed.

\subsection{Explanation  From Geometry Point of View}
We  provide some explanations on the above findings from the information geometry viewpoint. We note that  along with the training process, the model is  doing maximum likelihood estimation approximately by learning to fit the label distribution over the training data. It thus can be viewed as  a process that the \emph{log-likelihood landscape}  of the model on training data is gradually transiting  into a state where the model could  distinguish data of different categories well. On the other hand, the training data is   very sparsely sampled from the whole distribution. Therefore, during the formation of the model's log-likelihood landscape,  smoothness prior does not hold. Without such property, the model could easily use lower order local geometric descriptor such as Fisher information\textemdash the local curvature of the log-likelihood landscape\textemdash to form an overly simplified log-likelihood landscape that is adversary-prone due to lacking of smoothness. When applying adversarial training,   a strong smoothness property is enforced  and the model would have to rely on higher-order global geometric descriptor that vanishes locally to form the whole landscape. Thus the landscape could  be more robust to adversarial samples. 

\section{Towards Simultaneous Good  Performance and  Robustness}
With the disentanglement introduced above, it is natural to think whether it is possible to achieve decent standard accuracy and adversarial robustness simultaneously. From our results, if relying on the robust component alone can effectively distinguish data of different categories, obtaining an adversarial robust model with high standard accuracy is possible.

On the other hand, the expansion terms in the robust component are all higher order terms. Therefore, we suggest that the key to achieve the two desired qualities simultaneously is to increase the expressive power of model so that the model would have the ability to utilize the higher order terms for prediction. In this way, the model wouldn't have to rely heavily on Fisher information while still have decent standard performance with higher order terms (the robust component). Our disentanglement provides theoretical justification about the importance of model complexity in achieving adversarial robust and  relatively decent standard performance.

We also designed experiments on CIFAR-10 to provide insights into this possible strategy of achieving the two objective simultaneously. We train the following models with TRADES algorithm \cite{zhang2019theoretically}: a pure linear VGG13 model without ReLU  and with average pooling, a half linear VGG13 model without ReLU but using max pooling, a normal VGG 13 model, a normal resnet20 model and a resnet32 model with 10$\times$ more  channels. We evaluate these model on standard samples and adversarial samples produced by $L_{\infty}$ PGD attack \cite{madry2017towards} and CW attack \cite{carlini2017towards}. The results and experimental details are provided in Table \ref{adv-train-ablation}. 

We first compare VGG models. Pure linear VGG13 model is not complex enough to exploit higher order information for decision making during adversarial training. Thus, it cannot effectively leverage the lower order terms for prediction and achieves very poor accuracy on both standard and adversarial samples. However, as the non-linearity of the model increase (the half linear  and normal VGG13 models), the performance on standard and adversarial samples improves simultaneously. For resnet models, when the model is shallower and narrower (resnet20x1), the performance in both standard and adversarial settings are relatively low. However, with a deeper and wider resnet model (resnet32x10), the model could explore more higher order information for prediction, so the performance on standard samples and adversarial samples increases significantly.

  
    

\begin{table}
  \centering
    \caption{The standard accuracy and accuracy on adversary of models of different complexity trained by \cite{zhang2019theoretically}. We set the maximum allowed $L_{\infty}$ norm of attack noise $\epsilon = 8/255$. The learning rate 0.01 for all VGG13 models and 0.1 for all resnet models. All models are trained with SGD for 160 epochs with a decay of 10$\times$ in learning rate at 80 and 120 epochs. The SGD's momentum is 0.9. For adversarial training settings: the coefficient for the regularization term in \cite{zhang2019theoretically} is $\frac{1}{\lambda} = 5.0$, the step size for projected gradient descent is 2/255 and the number of step is 10. The weight decay during training is 1e-4. For evaluation against adversarial attack, the step size for PGD attack is 2/255, the number of step is 20. The CW attack objective's coefficient is $c=5e2$ and is also solved by PGD with the same optimization parameters. All experiments are conducted on NVIDIA Tesla V100 GPUs.}
  \begin{tabular}{cccccc}
    \toprule
    \cmidrule(r){1-2}
         & VGG13  & VGG13 & VGG13 &\multirow{2}{*}{resnet20x1} &\multirow{2}{*}{resnet32x10} \\
         & pure linear   &half linear  &normal & & \\
    \midrule
    accuracy  & \multirow{2}{*}{35.14} & \multirow{2}{*}{70.48} & \multirow{2}{*}{72.36} & \multirow{2}{*}{72.88} & \multirow{2}{*}{79.45} \\
     (standard) & & & & \\
    \midrule
    accuracy & \multirow{2}{*}{17.06}  & \multirow{2}{*}{38.52} & \multirow{2}{*}{42.29} & \multirow{2}{*}{44.94} & \multirow{2}{*}{49.52} \\ 
    (PGD attack,
    $\epsilon=8/255$)  &   &   &   &   &    \\
    \midrule
    accuracy & \multirow{2}{*}{13.94}  & \multirow{2}{*}{34.12} & \multirow{2}{*}{38.96} & \multirow{2}{*}{41.49} & \multirow{2}{*}{47.67} \\ 
    (CW attack,
    $\epsilon=8/255$)  &   &   &   &   &    \\
    \bottomrule
    \label{adv-train-ablation}
  \end{tabular}
\end{table}

\section{Discussion}
    \paragraph{When $\epsilon$ is NOT sufficiently small} Empirically, researchers have found that the local linearity assumption doesn't hold when allowed norm of attack noise $\epsilon$ is relatively larger \cite{kurakin2016adversarial,madry2017towards}. 
    In this case, our analysis is not so precise since higher order terms might contribute a lot in equation \eqref{eq: 12}. However, a higher order analogue of \eqref{eq: 12} and Fisher information could be available and explain the generalization behavior and adversarial robustness. For more details on this issue, we refer the reader to section B of appendix.
    \paragraph{Future directions} As we discussed before, our work has very strong geometry insight. Therefore, we suggest that it is possible that analyzing the geometric properties of the log-likelihood landscape formed by the DNNs on input data could provide us with even more interesting insights. Also, some recent literature \cite{weinan2017proposal,lu2017beyond,chen2018neural,zhang2019you} propose to view the deep learning model as a non-linear dynamic system and study it from the control point of view. The reachable state-space and the stability of the dynamic system are both widely studied in control theory, and we think that is corresponded to the performance and robustness to input perturbation in deep learning system. Also, when the reachable state-space of the dynamic system is large, it might be highly sensitive in certain directions in the state-space, which could also lead to the chaotic behavior of the dynamic system. This kind of trade off is very similar to the one that we discussed in our work. Therefore, we think it might also be possible to characterize the relationship between standard accuracy and adversarial behavior of deep learning models from dynamic system point of view.

\section{Conclusion}
In this work, we provide a novel view point on standard accuracy and adversarial robustness of deep learning model and show that the overall performance objective could be disentangled into a non-robust component, which is adversary-prone as well as a robust component, which is unrelated to adversarial behavior. In this way, we theoretically explain the relationship between standard accuracy and adversarial robustness: the cost of being adversarial robust is that the model could no longer effectively rely on the non-robust component to distinguish input data from different categories, which means that there is indeed a trade off between these two objectives. However, these two objectives might not be completely contradictory to each other, as there might exist a perfect balance point where the robust component of the model could perfectly distinguish input data from different categories while the non-robust component is not large enough to cause severe adversarial behavior. We also discussed the scenario where the norm of the allowed perturbation is not sufficiently small and higher order terms are needed in the expansion of adversarial objective and showed that our theory still holds. We're confident that more interesting theory about standard accuracy and adversarial robustness could be develop in the future based on our theory.


\medskip
\clearpage

\bibliographystyle{plain}
\bibliography{AderseralRobust}

\begin{thebibliography}{10}

\bibitem{achille2019task2vec}
Alessandro Achille, Michael Lam, Rahul Tewari, Avinash Ravichandran, Subhransu
  Maji, Charless Fowlkes, Stefano Soatto, and Pietro Perona.
\newblock Task2vec: Task embedding for meta-learning.
\newblock {\em arXiv preprint arXiv:1902.03545}, 2019.

\bibitem{andrychowicz2016learning}
Marcin Andrychowicz, Misha Denil, Sergio Gomez, Matthew~W Hoffman, David Pfau,
  Tom Schaul, Brendan Shillingford, and Nando De~Freitas.
\newblock Learning to learn by gradient descent by gradient descent.
\newblock In {\em Advances in Neural Information Processing Systems}, pages
  3981--3989, 2016.

\bibitem{carlini2017towards}
Nicholas Carlini and David Wagner.
\newblock Towards evaluating the robustness of neural networks.
\newblock In {\em 2017 IEEE Symposium on Security and Privacy (SP)}, pages
  39--57. IEEE, 2017.

\bibitem{chaudhry2018riemannian}
Arslan Chaudhry, Puneet~K Dokania, Thalaiyasingam Ajanthan, and Philip~HS Torr.
\newblock Riemannian walk for incremental learning: Understanding forgetting
  and intransigence.
\newblock In {\em Proceedings of the European Conference on Computer Vision
  (ECCV)}, pages 532--547, 2018.

\bibitem{chen2018neural}
Tian~Qi Chen, Yulia Rubanova, Jesse Bettencourt, and David~K Duvenaud.
\newblock Neural ordinary differential equations.
\newblock In {\em Advances in Neural Information Processing Systems}, pages
  6571--6583, 2018.

\bibitem{desjardins2015natural}
Guillaume Desjardins, Karen Simonyan, Razvan Pascanu, et~al.
\newblock Natural neural networks.
\newblock In {\em Advances in Neural Information Processing Systems}, pages
  2071--2079, 2015.

\bibitem{goodfellow2014explaining}
Ian~J Goodfellow, Jonathon Shlens, and Christian Szegedy.
\newblock Explaining and harnessing adversarial examples.
\newblock {\em arXiv preprint arXiv:1412.6572}, 2014.

\bibitem{he2015delving}
Kaiming He, Xiangyu Zhang, Shaoqing Ren, and Jian Sun.
\newblock Delving deep into rectifiers: Surpassing human-level performance on
  imagenet classification.
\newblock In {\em Proceedings of the IEEE international conference on computer
  vision}, pages 1026--1034, 2015.

\bibitem{he2016deep}
Kaiming He, Xiangyu Zhang, Shaoqing Ren, and Jian Sun.
\newblock Deep residual learning for image recognition.
\newblock In {\em Proceedings of the IEEE conference on computer vision and
  pattern recognition}, pages 770--778, 2016.

\bibitem{adversarialfeatures2019}
Andrew Ilyas, Shibani Santurkar, Dimitris Tsipras, Logan Engstrom, Brandon
  Tran, and Aleksander Madry.
\newblock Adversarial examples are not bugs, they are features.
\newblock {\em arXiv:1905.02175}, 2019.

\bibitem{ioffe2015batch}
Sergey Ioffe and Christian Szegedy.
\newblock Batch normalization: Accelerating deep network training by reducing
  internal covariate shift.
\newblock {\em arXiv preprint arXiv:1502.03167}, 2015.

\bibitem{krizhevsky2012imagenet}
Alex Krizhevsky, Ilya Sutskever, and Geoffrey~E Hinton.
\newblock Imagenet classification with deep convolutional neural networks.
\newblock In {\em Advances in neural information processing systems}, pages
  1097--1105, 2012.

\bibitem{kurakin2016adversarial}
Alexey Kurakin, Ian Goodfellow, and Samy Bengio.
\newblock Adversarial machine learning at scale.
\newblock {\em arXiv preprint arXiv:1611.01236}, 2016.

\bibitem{lecun2015deep}
Yann LeCun, Yoshua Bengio, and Geoffrey Hinton.
\newblock Deep learning.
\newblock {\em nature}, 521(7553):436, 2015.

\bibitem{lin1991divergence}
Jianhua Lin.
\newblock Divergence measures based on the shannon entropy.
\newblock {\em IEEE Transactions on Information theory}, 37(1):145--151, 1991.

\bibitem{lu2017beyond}
Yiping Lu, Aoxiao Zhong, Quanzheng Li, and Bin Dong.
\newblock Beyond finite layer neural networks: Bridging deep architectures and
  numerical differential equations.
\newblock {\em arXiv preprint arXiv:1710.10121}, 2017.

\bibitem{madry2017towards}
Aleksander Madry, Aleksandar Makelov, Ludwig Schmidt, Dimitris Tsipras, and
  Adrian Vladu.
\newblock Towards deep learning models resistant to adversarial attacks.
\newblock {\em arXiv preprint arXiv:1706.06083}, 2017.

\bibitem{miyato2015distributional}
Takeru Miyato, Shin-ichi Maeda, Masanori Koyama, Ken Nakae, and Shin Ishii.
\newblock Distributional smoothing with virtual adversarial training.
\newblock {\em arXiv preprint arXiv:1507.00677}, 2015.

\bibitem{moosavi2016deepfool}
Seyed-Mohsen Moosavi-Dezfooli, Alhussein Fawzi, and Pascal Frossard.
\newblock Deepfool: a simple and accurate method to fool deep neural networks.
\newblock In {\em Proceedings of the IEEE conference on computer vision and
  pattern recognition}, pages 2574--2582, 2016.

\bibitem{nakkiran2019adversarial}
Preetum Nakkiran.
\newblock Adversarial robustness may be at odds with simplicity.
\newblock {\em arXiv preprint arXiv:1901.00532}, 2019.

\bibitem{papernot2016limitations}
Nicolas Papernot, Patrick McDaniel, Somesh Jha, Matt Fredrikson, Z~Berkay
  Celik, and Ananthram Swami.
\newblock The limitations of deep learning in adversarial settings.
\newblock In {\em 2016 IEEE European Symposium on Security and Privacy
  (EuroS\&P)}, pages 372--387. IEEE, 2016.

\bibitem{papernot2016distillation}
Nicolas Papernot, Patrick McDaniel, Xi~Wu, Somesh Jha, and Ananthram Swami.
\newblock Distillation as a defense to adversarial perturbations against deep
  neural networks.
\newblock In {\em 2016 IEEE Symposium on Security and Privacy (SP)}, pages
  582--597. IEEE, 2016.

\bibitem{pascanu2013revisiting}
Razvan Pascanu and Yoshua Bengio.
\newblock Revisiting natural gradient for deep networks.
\newblock {\em arXiv preprint arXiv:1301.3584}, 2013.

\bibitem{schmidt2018adversarially}
Ludwig Schmidt, Shibani Santurkar, Dimitris Tsipras, Kunal Talwar, and
  Aleksander Madry.
\newblock Adversarially robust generalization requires more data.
\newblock In {\em Advances in Neural Information Processing Systems}, pages
  5014--5026, 2018.

\bibitem{simonyan2014very}
Karen Simonyan and Andrew Zisserman.
\newblock Very deep convolutional networks for large-scale image recognition.
\newblock {\em arXiv preprint arXiv:1409.1556}, 2014.

\bibitem{stutz2018disentangling}
David Stutz, Matthias Hein, and Bernt Schiele.
\newblock Disentangling adversarial robustness and generalization.
\newblock {\em arXiv preprint arXiv:1812.00740}, 2018.

\bibitem{szegedy2015going}
Christian Szegedy, Wei Liu, Yangqing Jia, Pierre Sermanet, Scott Reed, Dragomir
  Anguelov, Dumitru Erhan, Vincent Vanhoucke, and Andrew Rabinovich.
\newblock Going deeper with convolutions.
\newblock In {\em Proceedings of the IEEE conference on computer vision and
  pattern recognition}, pages 1--9, 2015.

\bibitem{szegedy2013intriguing}
Christian Szegedy, Wojciech Zaremba, Ilya Sutskever, Joan Bruna, Dumitru Erhan,
  Ian Goodfellow, and Rob Fergus.
\newblock Intriguing properties of neural networks.
\newblock {\em arXiv preprint arXiv:1312.6199}, 2013.

\bibitem{tsipras2018robustness}
Dimitris Tsipras, Shibani Santurkar, Logan Engstrom, Alexander Turner, and
  Aleksander Madry.
\newblock Robustness may be at odds with accuracy.
\newblock {\em stat}, 1050:11, 2018.

\bibitem{weinan2017proposal}
E~Weinan.
\newblock A proposal on machine learning via dynamical systems.
\newblock {\em Communications in Mathematics and Statistics}, 5(1):1--11, 2017.

\bibitem{zhang2019you}
Dinghuai Zhang, Tianyuan Zhang, Yiping Lu, Zhanxing Zhu, and Bin Dong.
\newblock You only propagate once: Painless adversarial training using maximal
  principle.
\newblock {\em arXiv preprint arXiv:1905.00877}, 2019.

\bibitem{zhang2019theoretically}
Hongyang Zhang, Yaodong Yu, Jiantao Jiao, Eric~P Xing, Laurent~El Ghaoui, and
  Michael~I Jordan.
\newblock Theoretically principled trade-off between robustness and accuracy.
\newblock {\em arXiv preprint arXiv:1901.08573}, 2019.

\bibitem{zhao2018adversarial}
Chenxiao Zhao, P~Thomas Fletcher, Mixue Yu, Yaxin Peng, Guixu Zhang, and
  Chaomin Shen.
\newblock The adversarial attack and detection under the fisher information
  metric.
\newblock {\em arXiv preprint arXiv:1810.03806}, 2018.

\end{thebibliography}

\clearpage

\appendix
\title{Appendix}
\maketitle
\section{More Experiments on the Role of Fisher Information}
\begin{figure}[h]
  \centering
  \includegraphics[width=10cm]{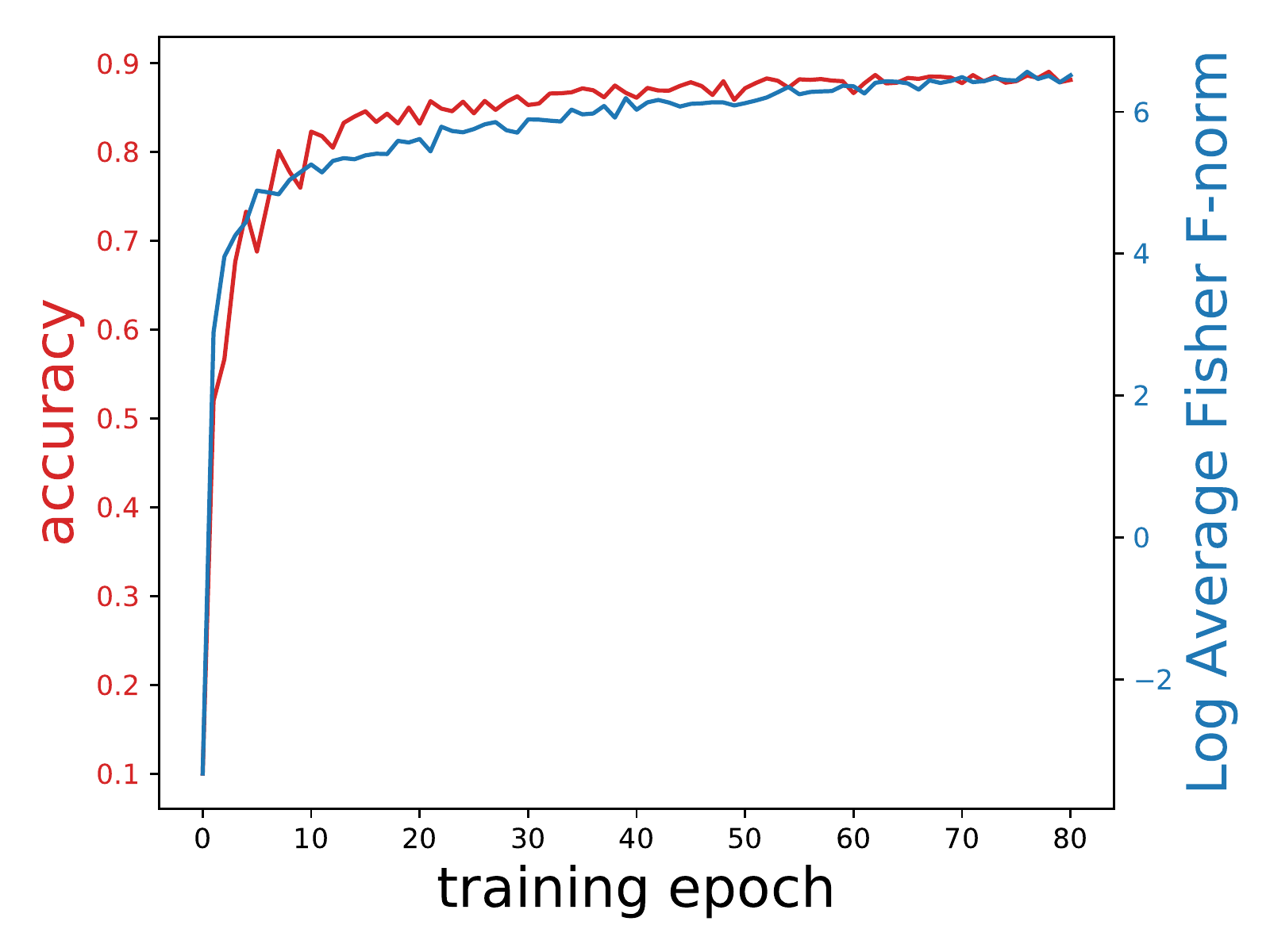}
  \caption{Visualization of how standard accuracy and average F-norm of Fisher information matrix on test set vary during training. The experiment is conducted on a resnet20 model and CIFAR-10 data set. We take the logarithm to better visualize the average F-norm of Fisher information}
  \label{fisher_resnet}
\end{figure}

\begin{figure}[h]
  \centering
  \includegraphics[width=6cm]{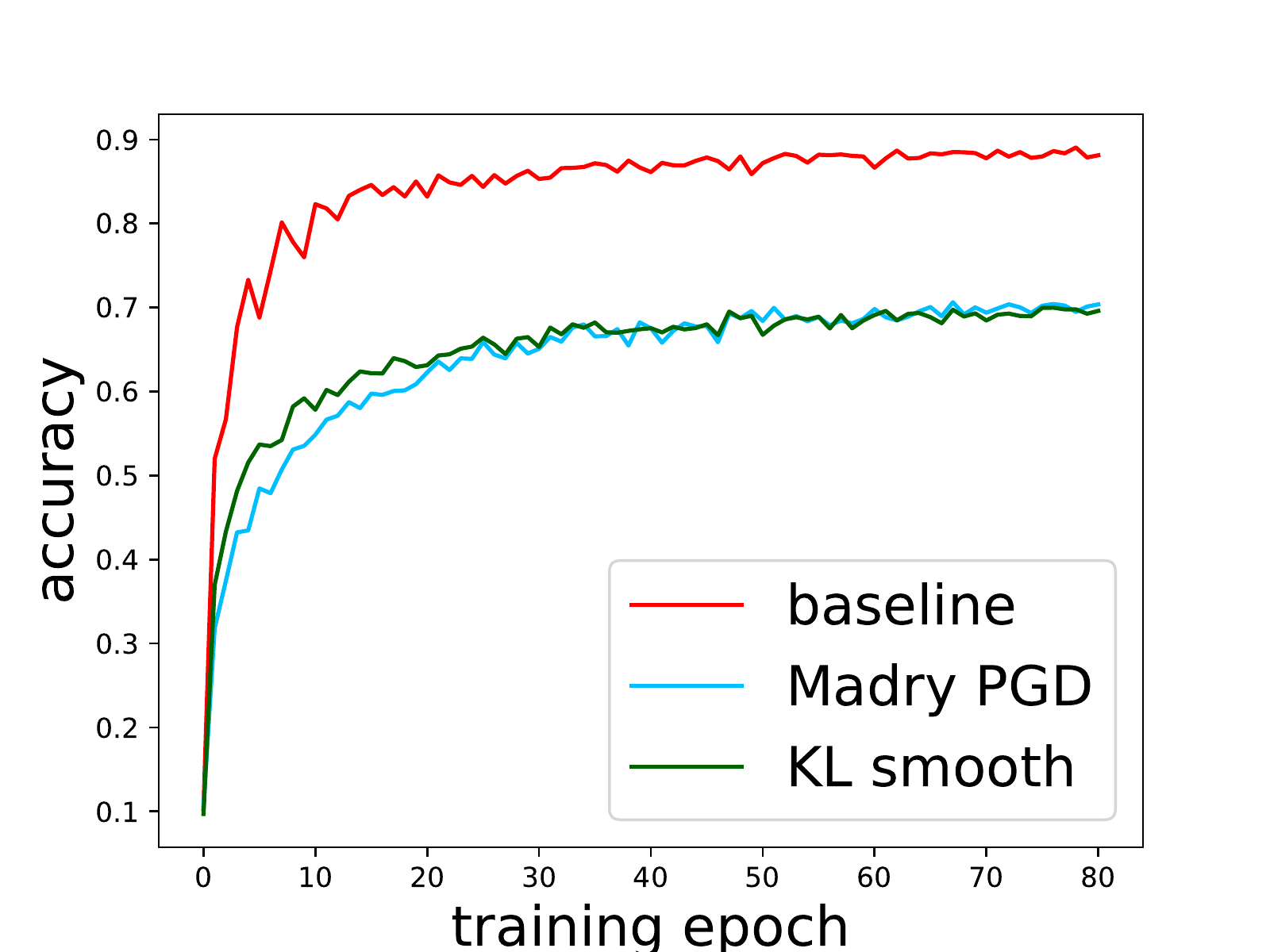}
  \includegraphics[width=6cm]{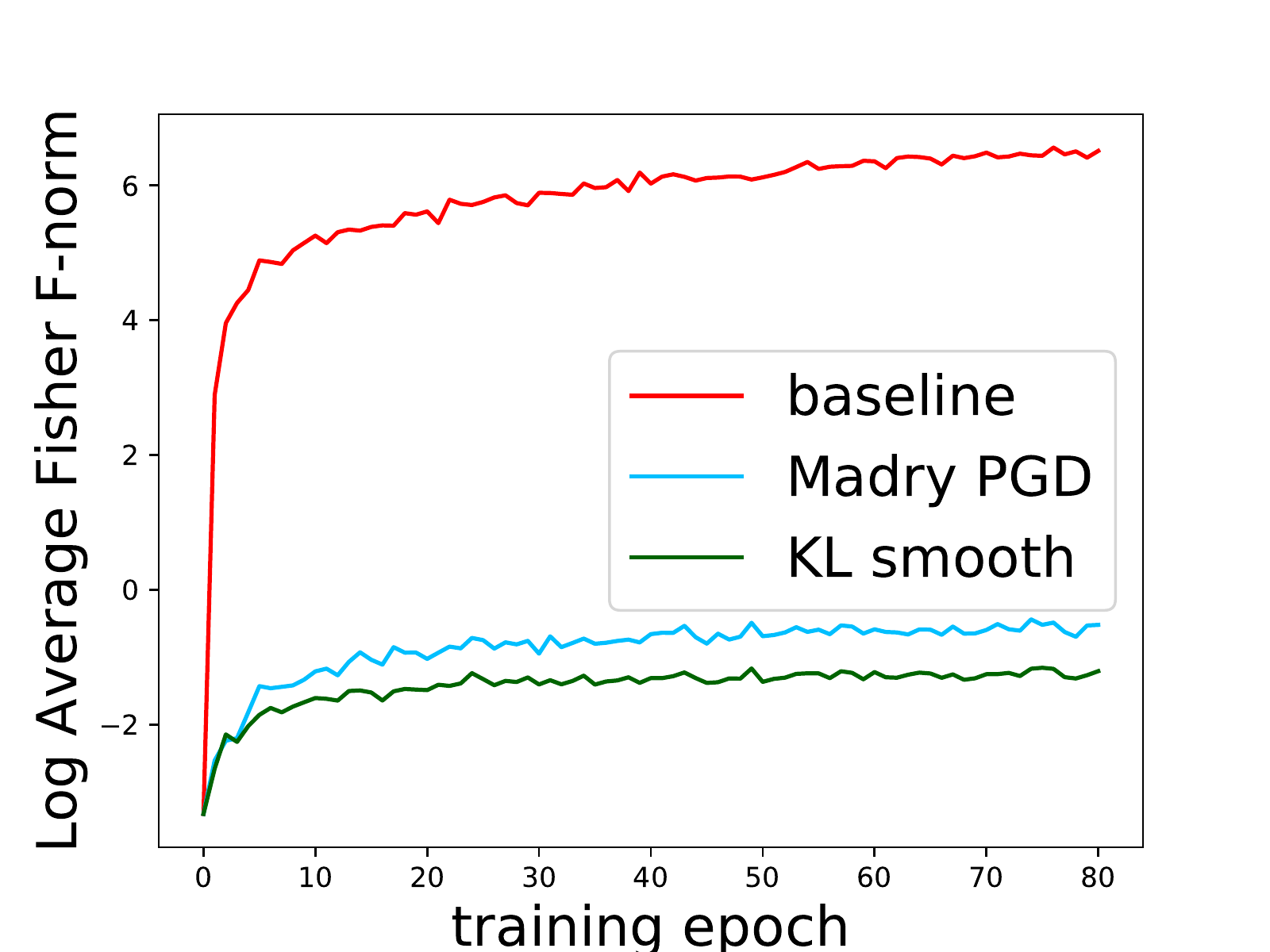}
  \caption{Visualization of how (left) standard test accuracy and (right) average F-norm of Fisher Information Matrix on test set vary during nature training and adversarial training, with resnet20 on CIFAR-10.}
  \label{fisher_resnet_adv}
\end{figure}

We conduct more visualization experiments about the role of Fisher information in standard performance of DNN. The results are shown in Figure \ref{fisher_resnet} and Figure \ref{fisher_resnet_adv}. The experiments are conducted on a resnet20 model. The same conclusion could be drawn according to our statistics.

\section{Adversarial behavior in large perturbation region}
    We now further discuss the scenario where the norm of adversarial noise is too large that the Fisher information alone is not enough to characterize the adversarial behavior of the model. However, since $\epsilon$ is still a small value, the expansion terms should still vanish at certain order $K_{\epsilon}$, where $K_{\epsilon}$ is an integer related to $\epsilon$. Therefore, for two data-label pairs $(x_i, y_i)$ and $(x_j, y_j)$, where $y_i \neq y_j$, the adversarial objective around $x_i$ should be rewritten as:
    \begin{equation}
        \label{eq: sup14}
        \max_{\eta} L_\theta (x_i, \eta) = \max_{\eta} \eta^\top F_{x_i}\eta + \sum_{k=3}^{K_{\epsilon}} T_{x_i}^{(k)}(\eta) \;\; \text{s.t.}\; \|\eta\| \leq \epsilon.
    \end{equation}
    We then rewrite the performance objective for better view:
    \begin{equation}
        \label{eq: sup15}
        L(x_i, x_j-x_i) = G_1 + G_2 =  \sum_{k=2}^{K_{\epsilon}} T_{x_i}^{(k)}(x_j-x_i) + \sum_{k=K_{\epsilon}+1}^{\infty} T_{x_i}^{(k)}(x_j-x_i).
    \end{equation}
    Here, we see that the adversarial behavior of the model around $x_i$ is not solely related to the spectral norm of $F_{x_i}$ any more\textemdash it is also related to the norm of the other $K_{\epsilon}{-}3$ multi-linear functional that yield the other $K_{\epsilon}{-}3$ terms. Similar to the derivation before, under the attack scale of $\epsilon$, if the model rely heavily on the first $G_1$ to distinguish input data from data belonging to other categories, then the norm of the first $K_{\epsilon}{-}2$ multi-linear functional in the expansion would have to increase drastically, and thus lead to more severe adversarial behavior around $x_i$ according to \eqref{eq: sup14}.

    Therefore, we show that even under the scenario where $\epsilon$ is not sufficiently small, the variant of our disentanglement could still clearly explain the relationship between achieving high standard accuracy and staying adversarial robust.

\end{document}